\documentclass[letterpaper]{article}
\usepackage{times}
\usepackage[stable]{footmisc}
\usepackage{graphicx}
\usepackage{breakcites}
\usepackage{algorithm,algorithmic,a4wide,amssymb,multicol,enumitem,url}
\usepackage{amsmath}
\usepackage{caption}
\usepackage{subcaption}
\usepackage[breaklinks]{hyperref}
 \usepackage[top=3.7cm, bottom=3.7cm, left=3.7cm, right=3.7cm]{geometry}

\usepackage{xcolor}

\makeatletter
\renewcommand\hyper@natlinkbreak[2]{#1}
\makeatother
\newcommand{\mytilde}{\raise.17ex\hbox{$\scriptstyle\mathtt{\sim}$}}
\begin{document}

\title{
Learning One Abstract Bit at a Time \\
Through Self-Invented Experiments \\ 
Encoded as Neural Networks
}

\author{%
    Vincent Herrmann$^1$ \\
    Louis Kirsch$^1$ \\
    J\"{u}rgen Schmidhuber$^{1,2,3}$ \vspace{0.2cm}\\
    $^1$\small{The Swiss AI Lab IDSIA/USI/SUPSI, Lugano, Switzerland} \\
    $^2$\small{NNAISENSE, Lugano, Switzerland} \\
    $^3$\small{AI Initiative, KAUST, Thuwal, Saudi Arabia} \\
    \small{\texttt{\{vincent.herrmann, louis, juergen\}@idsia.ch}}
}

\date{}
\maketitle

\begin{abstract}
\noindent There are two important things in science: (A) Finding answers to given questions, and (B) Coming up with good questions.
Our artificial scientists not only learn to answer given questions, but also continually invent new questions, by proposing hypotheses to be verified or falsified through potentially complex and time-consuming experiments, including thought experiments akin to those of mathematicians.
While an artificial scientist expands its knowledge, it remains biased towards the simplest, least costly experiments that still have surprising outcomes, until they become boring.
We present an empirical analysis of the automatic generation of interesting experiments.
In the first setting, we investigate self-invented experiments in a reinforcement-providing environment
and show that they lead to effective exploration.
In the second setting, pure thought experiments are implemented as the weights of recurrent neural networks generated by a neural experiment generator.
Initially interesting thought experiments may become boring over time. 
\end{abstract}

\section{Introduction \& Previous Work}
\label{intro}

We have frequently pointed out that there are two important things in science: (A) Finding answers to given questions, and (B) Coming up with good questions, e.g.,~\cite{Schmidhuber:90sab,powerplay2011and13,schmidhuber2020bit,schmidhuber2021cur}.
(A) is arguably just the standard problem of computer science. But how to implement the creative part (B) in artificial systems through reinforcement learning (RL), gradient-based artificial neural networks (NNs), and other machine learning methods? 

To answer this question,
for three decades we have published work on artificial scientists equipped with artificial curiosity and creativity, e.g.,~\cite{Schmidhuber:90diffenglish,Schmidhuber:90sab,Schmidhuber:91singaporecur,Storck:95,Schmidhuber:97interesting,Schmidhuber:06cs,Schmidhuber:10ieeetamd,sunyi2011agi,powerplay2011and13,Srivastava2013first,ramesh2022exploring}.
Our first artificial Q\&A system designed to invent and answer questions was the intrinsic motivation-based {\bf adversarial system} from 1990~\cite{Schmidhuber:90diffenglish,Schmidhuber:90sab}. 
It uses two artificial NNs. 
The first NN is called the controller $C$. 
$C$ probabilistically generates outputs that may influence an environment. 
The second NN is called the world model $M$. 
It predicts the environmental reactions to $C$'s outputs. 
Using gradient descent, M minimizes its error, thus becoming a better predictor. 
But in a zero-sum game, the reward-maximizing $C$ tries to find sequences of output actions that maximize the error of $M$. 
$M$'s loss is the gain of $C$ (like  in the later application of artificial curiosity  called GANs~\cite{schmidhuber2020gan}, but also for the more general cases of sequential data and RL~\cite{Kaelbling:96,Sutton:98,wiering2012}).

4 years before a 2014 paper on GANs~\cite{goodfellow2014generative}, a well-known 2010 survey \cite{Schmidhuber:10ieeetamd} summarised the generative adversarial NNs of 1990 as follows: a ``neural network as a predictive world model is used to maximize the controller's intrinsic reward, which is proportional to the model's prediction errors'' (which are minimized).
The 2014 GANs are an instance of this where the trials are very short (like in bandit problems) and the environment simply returns 1 or 0 depending on whether the controller's (or generator's) output is in a given set \cite{schmidhuber2020gan, schmidhuber2021cur, schmidhuber2022integr}.

$C$ is asking questions through its action sequences: 
What happens if I do that? 
$M$ is learning to answer those questions. 
$C$ is motivated to come up with questions where $M$ does not yet know the answer and loses interest in questions with known answers. 

This was the start of a long series of papers on artificial curiosity and artificial scientists~\cite{Schmidhuber:10ieeetamd,Schmidhuber:04cur,Schmidhuber:12cur,schmidhuber2020mir}.
Not only the 1990s but also more recent years saw successful applications of this simple principle (and variants thereof) in sequential settings, e.g.,~\cite{Singh:05nips,Oudeyer:12intrinsic,pathak2017curiosity,burda2018curious}.
Q\&As help to understand the world which is necessary for planning~\cite{Schmidhuber:90sandiego,Schmidhuber:90diffenglish,Schmidhuber:90sab} and may boost external reward~\cite{Schmidhuber:91singaporecur,Schmidhuber:02predictable,Schmidhuber:04cur,Schmidhuber:12cur,pathak2017curiosity,burda2018curious}.

The approach of 1990~\cite{Schmidhuber:90diffenglish,Schmidhuber:90sab} makes for a fine exploration strategy in many deterministic environments.
{\bf In stochastic environments, however, it might fail.}
$C$ might learn to focus on those parts of the environment where $M$ can always get high prediction errors due to randomness, or due to computational limitations of $M$. 
For example, an agent controlled by $C$ might get stuck in front of a TV screen showing highly unpredictable white noise, e.g.,~\cite{Schmidhuber:10ieeetamd} (see also~\cite{burda2018curious}). 

Therefore, as pointed out in 1991, in stochastic environments, $C$'s reward  should not be the errors of $M$, but (an approximation of) the {\em first derivative} of $M$'s errors across subsequent training iterations,
that is, $M$'s {\bf learning progress or improvements}~\cite{Schmidhuber:91singaporecur,Schmidhuber:07alt}. 
As a consequence, despite $M$'s high errors in front of a noisy TV screen, $C$ won't get rewarded for getting stuck there, simply because $M$'s errors won't improve.
Both the totally predictable and the fundamentally unpredictable will get boring. 

This simple insight led to lots of follow-up work~\cite{Schmidhuber:10ieeetamd}. 
For example, one particular RL approach for artificial curiosity in stochastic environments was published in 1995~\cite{Storck:95}. 
A simple $M$ learned to predict or estimate the probabilities of the environment's possible responses, given $C$'s actions. 
After each interaction with the environment, $C$'s intrinsic reward was the KL-Divergence~\cite{kullback1951} between $M$'s estimated probability distributions
before and after the resulting new experience---the {\bf information gain}~\cite{Storck:95}.
This was later also called {\em Bayesian Surprise}~\cite{itti:05}.
Compare earlier work on information gain~\cite{Shannon:48} and its maximization {\em without} RL \& NNs~\cite{Fedorov:72}.

In the general RL setting where the environment is only partially observable~\cite[Sec.~6]{888}, $C$ and $M$ may greatly profit from a memory of previous events~\cite{Schmidhuber:90sandiego,Schmidhuber:90diffenglish,Schmidhuber:91nips}.
Towards this end, both $C$ and $M$ can be implemented as 
LSTMs~\cite{lstm97and95,Gers:2000nc,Graves:09tpami,888}
which have become highly commercial
~\cite{googlevoice2015,wu2016google,amazon2016,facebook2017} and widely used in RL~\cite{wierstra2010,openai2019dota,deepmind2019starcraft,openai2020dex}.

The better the predictions of $M$, the fewer bits are required to encode the history $H$ of observations because short codes can be used for observations that $M$ considers highly probable~\cite{Huffman:52,Witten:87}. 
That is, the learning progress of $M$ has a lot to do with the concept of {\em compression progress}~\cite{Schmidhuber:06cs,Schmidhuber:09abials,Schmidhuber:09videos,Schmidhuber:10ieeetamd}. 
But it's not quite the same thing. 
In particular, it does not take into account the bits of information needed to specify $M$.
A more general approach is based on algorithmic information theory, e.g.,~\cite{Solomonoff:64,Kolmogorov:65,Wallace:68,Wallace:87,LiVitanyi:97,Schmidhuber:02ijfcs}.
Here $C$'s intrinsic reward is indeed based on {\bf  algorithmic compression progress}~\cite{Schmidhuber:06cs,Schmidhuber:09abials,Schmidhuber:09videos,Schmidhuber:10ieeetamd} based on
some coding scheme for the weights of the model network, e.g.,~\cite{Hochreiter:97nc1,Schmidhuber:95kol+,Schmidhuber:97nn,koutnik:gecco10,ppsn2012cncs,koutnik:gecco13,steenkiste2016wavelet}, and also a coding scheme for the history of all observations so far, given the model~\cite{Huffman:52,Wallace:68,Rissanen:78,Witten:87,Hochreiter:97nc1,Schmidhuber:06cs}.
Note that the history of science is a history of compression progress through incremental discovery of simple laws that govern seemingly complex observation sequences~\cite{Schmidhuber:06cs,Schmidhuber:09abials,Schmidhuber:09videos,Schmidhuber:10ieeetamd}.

Back in 1990, the questions asked by $C$ were restricted in the sense that they always referred to all the details of future inputs, e.g., pixels~\cite{Schmidhuber:90diffenglish,Schmidhuber:90sab}. 
That’s why in 1997, a more general adversarial RL machine was built that could ignore many or all of these details and ask {\bf arbitrary abstract questions} with computable answers~\cite{Schmidhuber:97interesting,Schmidhuber:99cec,Schmidhuber:02predictable}. 
Example question: if we run this policy (or program) for a while until it executes a special interrupt action, will the internal storage cell number 15 contain the value 5, or not? Again there are two learning, reward-maximising adversaries playing a zero-sum game, occasionally betting on different yes/no outcomes of such computational experiments. 
The winner of such a bet gets a reward of 1, the loser -1.  
So each adversary is motivated to come up with questions whose answers surprise the other. 
And both are motivated to avoid seemingly trivial questions where both already agree on the outcome, or seemingly hard questions that none of them can reliably answer for now. 
This is the approach closest to what we will present in the following sections.

All the systems above (now often called CM systems~\cite{learningtothink2015}) actually maximize the sum of the standard external rewards (for achieving user-given goals) and the intrinsic rewards. 
{\bf Does this distort the basic RL problem?} 

It turns out not so much. Unlike the external reward for eating three times a day, the curiosity reward in the systems above is ephemeral, because once something is known, there is no additional intrinsic reward for discovering it again.
That is, the external reward tends to dominate the total reward. In totally learnable environments, in the long run, the intrinsic reward even  {\em vanishes} next to the external reward. 
Which is nice, because in most RL applications we care only for the external reward. 

Our RL Q\&A systems of the 1990s did not {\bf explicitly, formally enumerate their questions.} 
But the more recent  {\sc PowerPlay} framework (2011)~\cite{powerplay2011and13,Srivastava2013first} does. 
Let us step back for a moment. 
What is the set of all formalisable questions? 
How to decide whether a given question has been answered by a learning machine? 
To define a question, we need a computational procedure that takes a solution candidate (possibly proposed by a policy) and decides whether it is an answer to the question or not. 
{\sc PowerPlay} essentially enumerates the set of all such procedures (or some user-defined subset thereof), thus enumerating all possible questions or problems. 
{\bf It searches for the simplest question that the current policy cannot yet answer but can quickly {\em learn}  to answer {\em without} forgetting the answers to previously answered questions.} 
What is the simplest such Q\&A to be added to the repertoire? 
It is the cheapest one---the one that is found first. 
Then the next trial starts, where new Q\&As may build on previous Q\&As.  
Compare also the {\em One Big Net For Everything}~\cite{onebignet2018} which offers a simplified, less strict NN version of {\sc PowerPlay}.

In our empirical investigation of Section~\ref{sec:empiricial}, we will revisit the above-mentioned concepts of complex computational experiments with yes/no outcomes, focusing on two settings:
(1) The generation of experiments driven by model prediction error in a deterministic reinforcement-providing environment,
and (2) An approach where $C$ (driven by information gain) generates pure thought experiments in form of weight matrices of RNNs.

\section{Self-Invented Experiments Encoded as Neural Networks}
\label{exabs}

We present a $CM$ system where $C$ can design essentially arbitrary computational experiments (including thought experiments) with binary yes/no outcomes.
Experiments may run for several time steps. 
However, $C$ will prefer simple experiments whose outcomes still surprise $M$, until they become boring.

In general, both the controller $C$ and the model $M$ can be implemented as (potentially multi-dimensional) LSTMs \cite{graves:icann2007}. 
At each time step  $t=1,2, \ldots$, $C$'s input includes the current sensory input vector $in(t)$, the external reward vector $R_e(t)$, and the intrinsic curiosity reward $R_i(t)$.
$C$ may or may not interact directly with the environment through action outputs. 
How does $C$ ask questions and propose experiments?
$C$ has an output unit called the START unit. Once it becomes active ($>0.5$), 
$C$ uses a set of extra output units for producing the {\em weight matrix or program} $\theta$ of a separate RNN or LSTM called $E$ (for Experiment), in fast weight programmer style~\cite{Schmidhuber:92ncfastweights,Schmidhuber:91singaporefastweights,Schmidhuber:93ratioicann,Gomez:05icann,schlag2018tensor, faccio2020parameter, kirsch2021meta, schlag2021linear, irie2021going}.

$E$ takes sensory inputs from the environment and produces actions as outputs.
It also has two additional output units, the HALT unit~\cite{Schmidhuber:12slimnn} and the RESULT unit.
Once the weights $\theta$ are generated at time step $t'$, $E$ is tested in a trial, interacting with some environment.
Once $E$'s HALT unit exceeds 0.5 in a later time step $t''$, 
the current experiment ends. That is, the experiment computes its own runtime~\cite{Schmidhuber:12slimnn}.
The experimental outcome $r(t'')$ is $1$ if the activation {\em result}$(t'')$ of $E$'s RESULT unit exceeds $0.5$, and $0$ otherwise. 
  
At time $t'$, so before the experiment is being executed, $M$ has to compute its output {\em pr}$(t') \in [0,1]$ from $\theta$ (and the history of $C$'s inputs and actions up to $t'$, which includes all previous experiments their outcomes).
Here, {\em pr}$(t')$ models $M$'s (un)certainty that the final binary outcome of the experiment will be 1 (YES) or 0 (NO).
Then the experiment is run.

In short, $C$ is proposing an experimental question in form of $\theta$ that will yield a binary answer (unless some time limit is reached).
$M$ is trying to predict this answer before the experiment is executed.
Since $E$ is an RNN and thus a general computer whose weight matrix can implement any program executable on a traditional computer~\cite{siegelmann91turing}, any computable experiment with a binary outcome can be implemented in its weight matrix (ignoring storage limitations of finite RNNs or other computers). 
That is, by generating an appropriate weight matrix $\theta$, $C$ can ask any scientific question with a computable solution. 
In other words, $C$ can propose any scientific hypothesis that is experimentally verifiable or falsifiable.

At  $t''$, $M$'s previous prediction  {\em pr}$(t')$ is compared to the later observed outcome  $r(t'')$ of C's experiment (which spans $t''-t'$ time steps), and $C$'s intrinsic curiosity reward $R_i(t'')$ is proportional to $M$'s surprise. 
To calculate it, we interpret  {\em pr}$(t')$ as $M$'s estimated probability of $r(t'')$, given the history of observations so far.
Then we train $M$ by gradient descent (with regularization to avoid overfitting) for a fixed amount of time to improve all of its previous predictions including the most recent one.
This yields an updated version of $M$ called $M^*$.

In general, $M^*$ will compute a different prediction  {\em PR}$(t')$ of $r(t'')$, given the history up to $t'-1$. 
At time $t''$, the contribution $R_{IG}(t'')$ to $C$'s curiosity reward is proportional to the apparent resulting information gain, the KL-divergence

\[
R_{IG}(t'') \sim D_{KL} \big(PR(t') || pr(t')\big).
\]

If $M$ had a confident belief in a particular experimental outcome, but this belief gets shattered in the wake of $C$'s experiment, there will be a major surprise and a big insight for $M$, as well as lots of intrinsic curiosity reward for $C$. 
On the other hand, if $M$ was quite unsure about the experimental outcome, and remains quite unsure afterwards, then $C$'s experiment can hardly surprise $M$ and $C$ will fail to profit much. 
$C$ is motivated to propose  {\em interesting} hypotheses or experiments that violate $M$'s current deep beliefs and expand its horizon. 
An alternative intrinsic curiosity reward would be based on compression progress~\cite{Schmidhuber:06cs,Schmidhuber:09abials,Schmidhuber:09videos,Schmidhuber:10ieeetamd}.

Note that the entire experimental protocol is the responsibility of $\theta$.
Through $\theta$, $E$ must initialize the experiment (e.g., by resetting the environment or moving the agent to some start position if that is important to obtain reliable results), then run the experiment by executing a sequence of computational steps or actions, and translate the incoming data sequence into some final abstract binary outcome YES or NO. 

$C$ is motivated to design experimental protocols $\theta$ that surprise $M$. 
$C$ will get bored by experiments whose outcomes are predicted by $M$ with little confidence (recall the noisy TV), as well as by experiments whose outcomes are correctly predicted by $M$ with high confidence.
{\em $C$ will get rewarded for surprising experiments whose outcomes are incorrectly predicted by $M$ with high confidence.}

A negative reward per time step encourages $C$ to be efficient and lazy and come up with simple and fast still surprising experiments. 
If physical actions in the environment cost much more energy (resulting in immediate negative reward) than $E$'s internal computations per time step,
$C$ is motivated to propose a $\theta$ defining a ``thought experiment'' requiring only internal computations, without executing physical actions in the (typically non-differentiable) environment. 
In fact, due to $C$'s bias towards the computationally cheapest and least costly experiments that are still surprising to $M$, most of $C$'s initial experiments may be thought experiments. 
Hence, since $C$, $E$ and $M$ are differentiable, not only $M$ but also $C$ may be often trainable by backpropagation~\cite{faccio2020parameter} rather than the generally slower policy gradient methods~\cite{wierstra2010,openai2019dota,deepmind2019starcraft,openai2020dex}.
Of course, this is only true if the reward function is also differentiable with respect to $C$'s parameters.

\section{Experimental Evaluation}
\label{sec:empiricial}

Here we present initial studies of the automatic generation of interesting experiments encoded as NNs. 
We evaluate these systems empirically and discuss the associated challenges. 
This includes two setups: (1) Adversarial intrinsic reward  encourages experiments executed in a differentiable  environment through sequences of continuous control actions.
We demonstrate that these experiments aid the discovery of goal states in a sparse reward setting. 
(2) Pure thought experiments encoded as RNNs (without any environmental interactions) are guided by an information gain reward.

Together, these two setups cover the important aspects discussed in Section~\ref{exabs}: 
the use of abstract experiments with binary outcomes as a method for curious exploration, and the creation of interesting pure thought experiments encoded as RNNs. 
We leave the integration of both setups into a single system (as described in section~\ref{exabs}) for future work.

\subsection{Generating Experiments in a Differentiable Environment}
\label{sec:differentiable_env}

Reinforcement learning (RL) usually involves exploration in an environment with non-differentiable dynamics.
This requires RL methods such as policy gradients~\cite{Williams:92}.
To simplify our investigation and focus solely on the generation of self-invented experiments, we introduce a fully differentiable environment that allows for computing analytical policy gradients via backpropagation.
This does not limit the generality of our approach, as standard RL methods can be used instead.

\begin{figure}[t]
     \centering
     \begin{subfigure}[t]{0.45\textwidth}
         \centering
         \includegraphics[width=\textwidth]{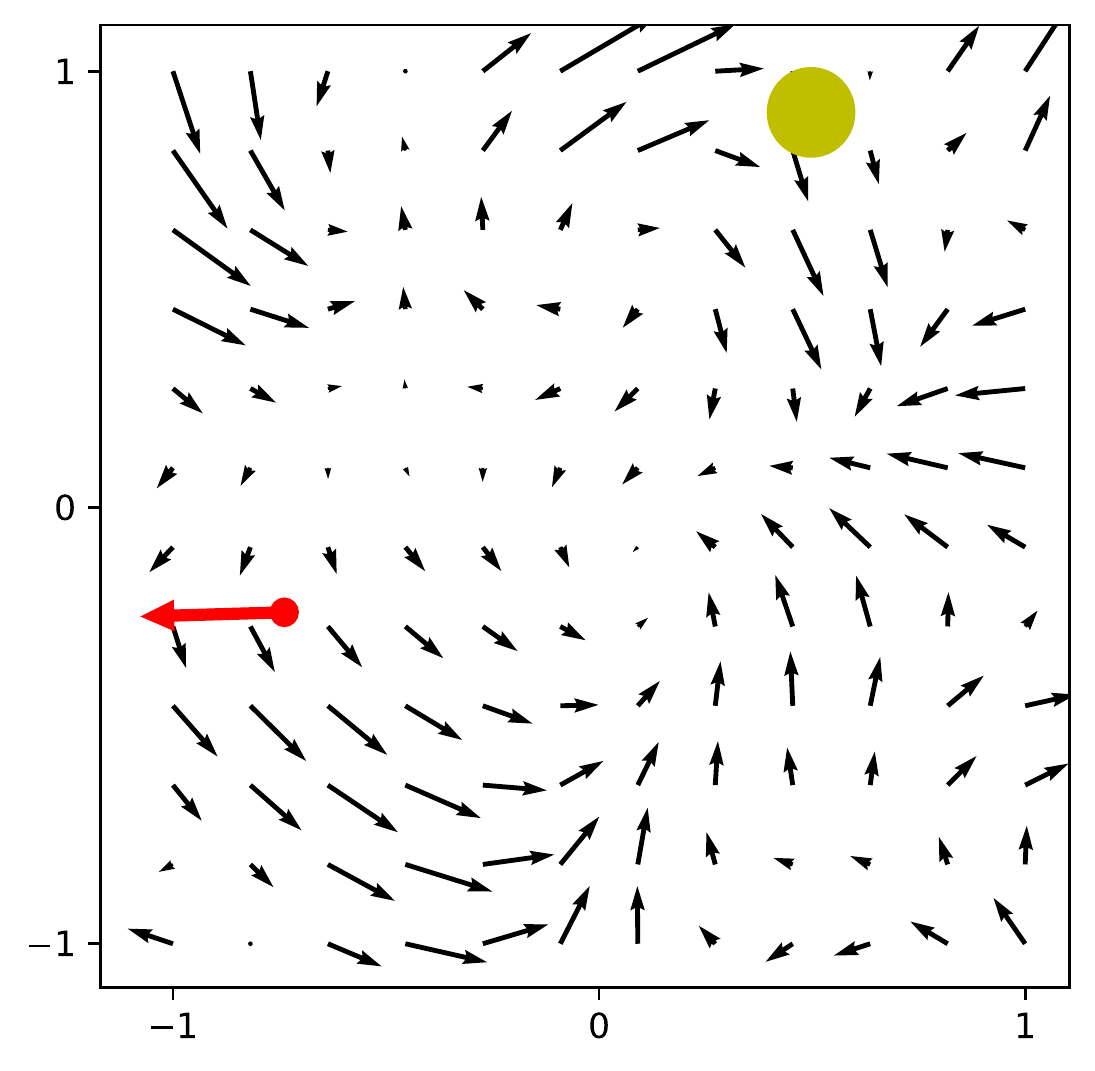}
     \end{subfigure}
     \caption{\textbf{A differentiable force field environment}. The agent (red) has to navigate to the goal state (yellow) while the external force field exerts forces on the agent.}
     \label{fig:environment}
     \hfill
\end{figure}

Our continuous force field environment is depicted in Figure~\ref{fig:environment}.
The agent has to navigate through a 2D environment with a fixed external force field.
This force field can have different levels of complexity.
The states in this environment are the position and velocity of the agent.
The agent's actions are real-valued force vectors applied to itself. To encourage laziness and a bias towards simple experiments,
each time step is associated with a small negative reward ($-0.1$).
A sparse large reward ($100$) is given whenever the agent gets very close to the goal state.
We operate in the single life setting without episodic resets.
Additional information about the force field environment can be found in Appendix~\ref{app:environment}.
Since the environment is deterministic, it is sufficient for $C$ to generate experiments whose results the current $M$ cannot predict.

\begin{figure}[ht]
     \centering
     \begin{subfigure}[t]{0.45\textwidth}
         \centering
         \includegraphics[width=\textwidth]{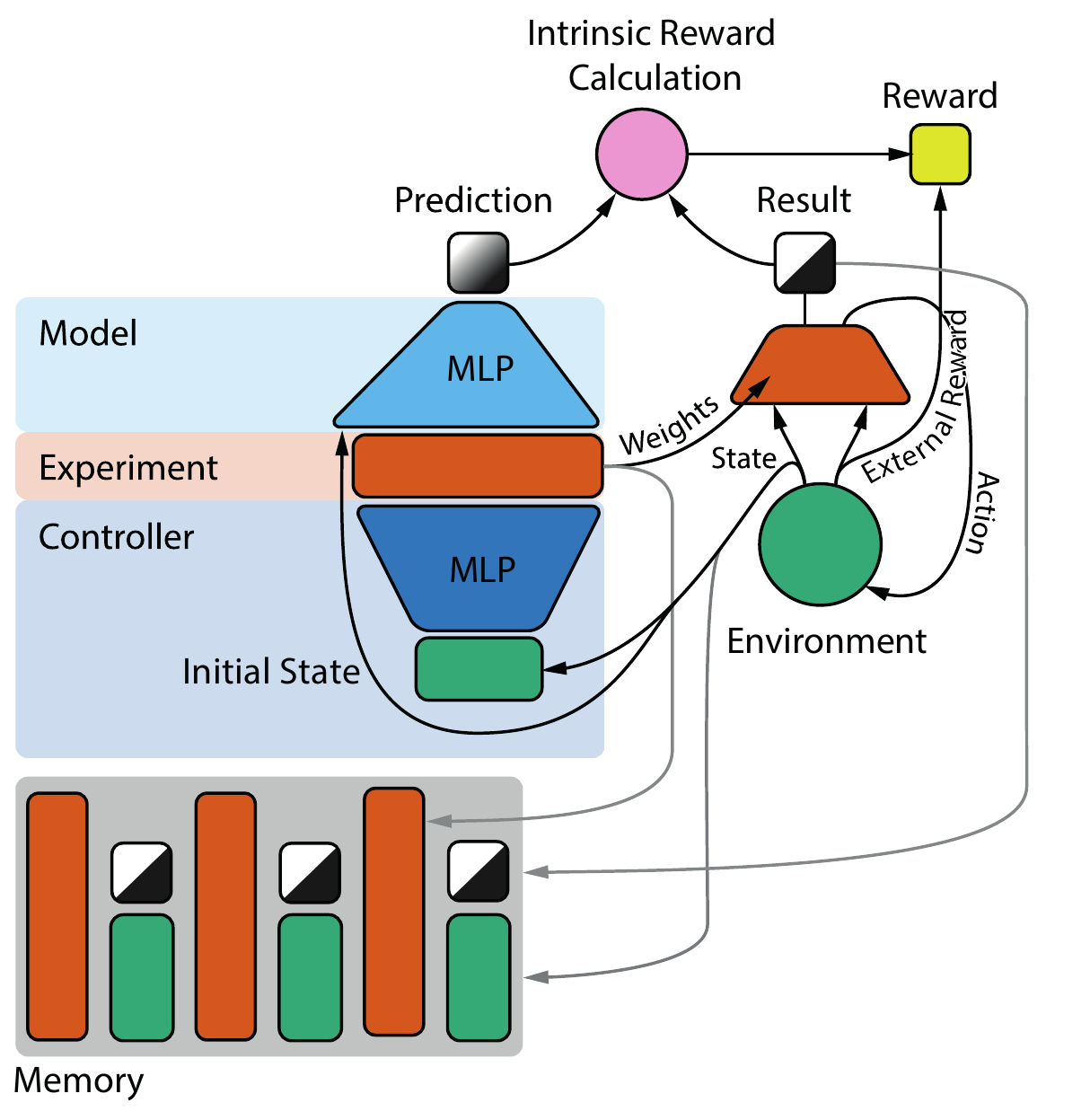}
     \end{subfigure}
     \caption{
     \textbf{Generating self-invented experiments in a differentiable environment.}
     A controller $C_\phi$ is motivated to generate experiments $E_\theta$ that still surprise the model $M_\mathbf{w}$.
     After execution in the environment, the experiments and their binary results are stored in memory. The model is trained on the history of previous experiments.
     }
     \label{fig:adversarial_env}
     \hfill
\end{figure}

\subsubsection{Method}

Algorithm~\ref{alg:adversarial} and Figure~\ref{fig:adversarial_env} summarize the process for generating a sequence of interesting abstract experiments with binary outcomes. 
The goal is to test the following three hypotheses:
\begin{itemize}
    \item Generated experiments implement exploratory behavior, facilitating the reaching of goal states.
    \item If there are negative rewards in proportion to the runtime of experiments, then the average runtime will increase over time, as the controller will find it harder and harder to come up with new short experiments whose outcomes the model cannot yet predict.
    \item As the model learns to predict the yes/no results of more and more experiments, it becomes harder for the controller to create experiments whose outcomes surprise the model.
\end{itemize}

\noindent The generated experiments have the form $E_\psi(s) = (a, \hat{r})$, where $E_\psi$ is a linear feedforward network with parameters $\psi$, $s$ is the environment state, $a$ are the actions and $\hat{r} \in [0, 1]$ is the experimental  result. 
Both $s$ and $a$ are real-valued vectors.

Instead of a HALT unit, a single scalar $\tau \in \mathbb{R}^+$ determines the number of steps for which an experiment will run.
To further simplify the setup, the experiment network is a feedforward NN without recurrence.
To make the experimental result differentiable with respect to the runtime parameter, $\tau$ predicts the mean of a Gaussian distribution with fixed variance over the number of steps.
The actual result $\tilde{r}$ is the expectation of the result unit $\hat{r}$ over the distribution defined by $\tau$ (more details on this can be found in Appendix~\ref{app:experiments}).
The binarized result $r$ has the value 1 if $\tilde{r} > 0.5$, and $0$ otherwise. 
The parameters $\theta$ of the experiment are the network parameters $\psi$ together with the runtime parameter $\tau$, i.e. $\theta := (\psi, \tau)$.

For a given starting state $s$, the controller $C_\phi$ generates experiments: $C_\phi(s) = \theta$.
$C_\phi$ is a multi-layer perceptron (MLP) with parameters $\phi$, and $\theta$ denotes the parameters of the generated experiment.
The model $M_\mathbf{w}$ is an MLP with parameters $\mathbf{w}$.
It makes a prediction $M_\mathbf{w}(s, \theta) = \hat{o}$, with $\hat{o} \in [0, 1]$, for an experiment defined by the starting state $s$ and the parameters $\theta$.

During each iteration of the algorithm, $C_\phi$ generates an experiment based on the current state $s$ of the environment. 
This experiment is executed until the cumulative halting probability defined by the generated $\tau$ exceeds a certain threshold (e.g., 99\%).
The starting state $s$, experiment parameters $\theta$ and binary result $r$ are saved in a memory buffer $\mathcal{D}$ of experiments.
Every state encountered during the experiment is saved to the state memory buffer $\mathcal{B}$.

After the experiment execution, the model $M_\mathbf{w}$ is trained for a fixed number of steps of stochastic gradient descent (SGD) to minimize the loss 
\begin{equation}
\label{eq:loss_m}
    \mathcal{L}_M = \mathbb{E}_{(s, \theta, r) \sim \mathcal{D}}[\text{bce}(M_\mathbf{w}(s, \theta), r)],
\end{equation} 
where $\text{bce}$ is the binary cross-entropy loss function.

The third and last part of each iteration is the training of the controller $C_\phi$.
The loss that is being minimized via SGD is 
\begin{equation}
\label{eq:loss_c}
    \mathcal{L}_C = \mathbb{E}_{s \sim \mathcal{B}} [- \text{bce}\big( M_\mathbf{w}(s, C_\phi(s)), \tilde{r}(C_\phi(s), s) \big) - R_e(C_\phi(s), s)].
\end{equation}

The function $\tilde{r}$ maps the experiment parameters and  starting state to the continuous result of the experiment.
The function $R_e$ maps the experiment parameters and starting state to the external reward.
Note that gradient information will flow back from $\tilde{r}$ and $R$ to $\phi$ through the execution of the experiment in the differentiable environment.
The first term corresponds to the intrinsic reward for the controller, which encourages it to generate experiments whose outcomes $M_\textbf{w}$ cannot predict.
The second term is the external reward from the environment, which punishes long experiments.
Since the reward for reaching the goal is sparse and not differentiable with respect to the experiment's actions, no information about the goal state reaches $C_\phi$ through the gradient.

\begin{figure}[t]
     \centering
     \begin{subfigure}[t]{0.44\textwidth}
         \centering
         \includegraphics[width=\textwidth]{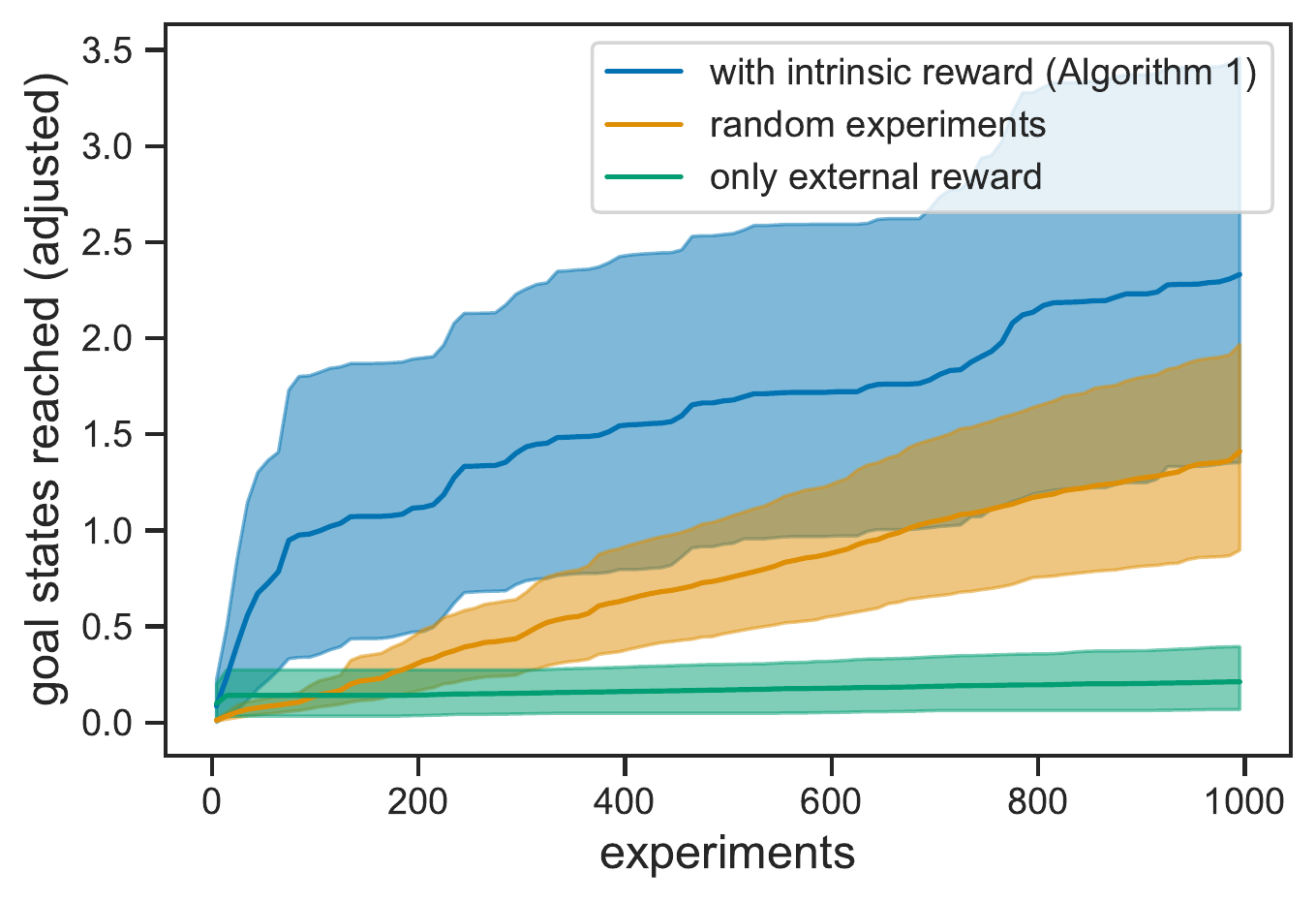}
         \caption{Number of times the goal state was reached, adjusted by the number of environment interactions. Experiments generated with adversarial intrinsic reward benefit exploration more  than random experiments. Without intrinsic motivation, the agent usually fails to reach any goal states in the sparse reward setting. Mean with bootstrapped 95\% confidence intervals across 30 seeds.}
         \label{fig:force_field_goal_states}
     \end{subfigure}
     \hfill
     \begin{subfigure}[t]{0.5\textwidth}
         \centering
         \includegraphics[width=\textwidth]{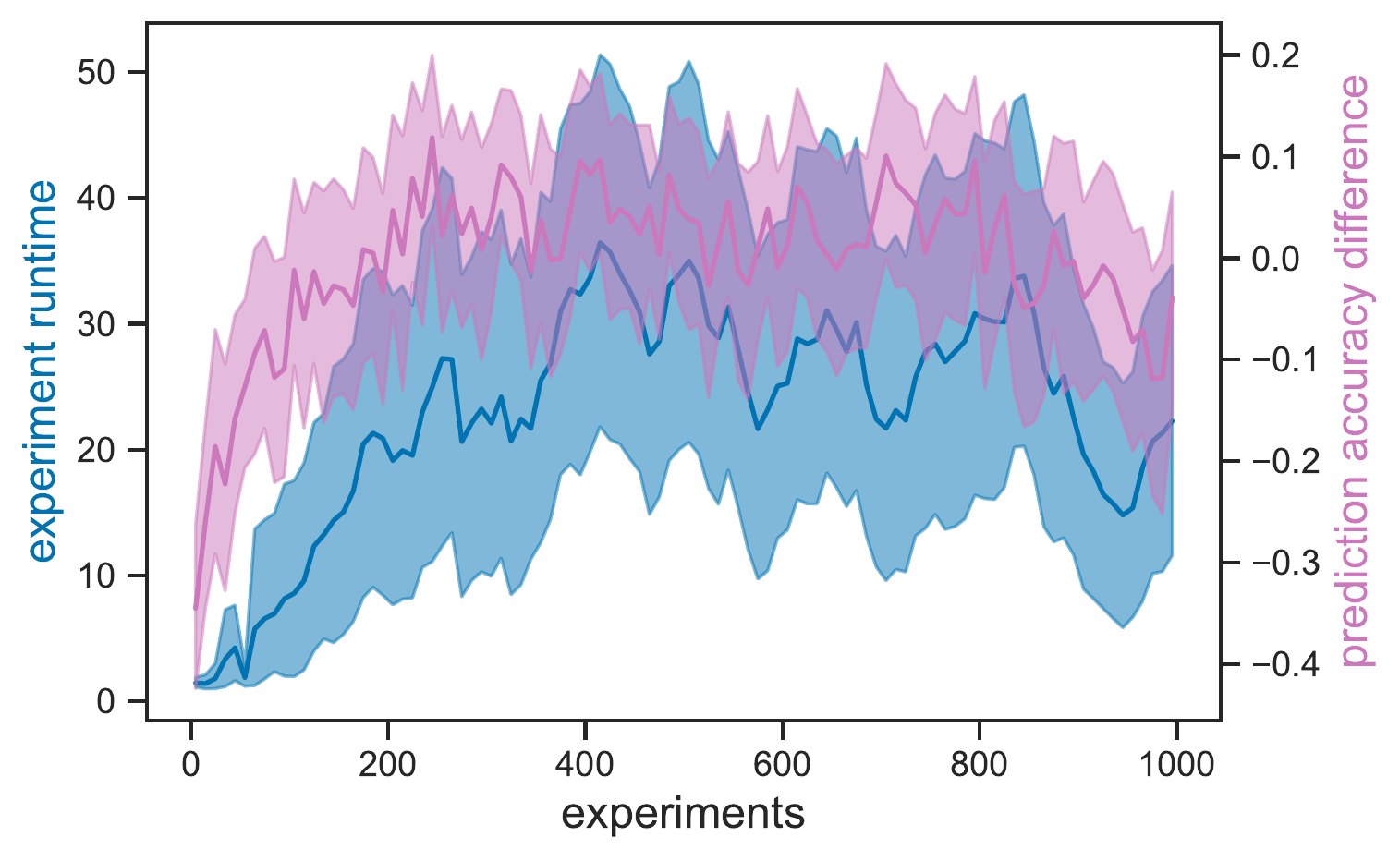}
         \caption{Blue: the average runtime of each experiment generated by $C\phi$. Purple: the difference between result prediction accuracy of the current $M_\mathbf{w}$ for the generated experiment and the average prediction accuracy of the current $M_\mathbf{w}$ for random experiments. Mean with bootstrapped 95\% confidence intervals across 30 seeds.}
         \label{fig:force_field_runtime}
     \end{subfigure}
     \label{fig:force_field_results}
     \caption{Experiments in the differentiable force field environment}
\end{figure}

\subsubsection{Results and Discussion}
\label{sec:force_results}

To investigate our first hypothesis, Figure~\ref{fig:force_field_goal_states} shows the cumulative number of times a goal state was reached during an experiment, adjusted by the number of environment interactions of each experiment.
Specifically, it shows $h(j) = \sum_{k=1}^j \frac{g_k}{n_k}$, where $j = 1, 2, \ldots$ is the index of the generated experiment, $g_k$ is $1$ if the goal state was reached during the $k$th experiment and $0$ otherwise, and $n_k$ is the runtime of the $k$th experiment.
Our method, as described above and in Algorithm~\ref{alg:adversarial}, reaches the most goal states per environment interaction.
Purely random experiments also discover goal states, but less frequently.
Note that such random exploration in parameter space has been shown to be a powerful exploration strategy \cite{rueckstiess2008b, plappert2017parameter, vemula2019contrasting}.
The average runtime of the random experiments is $50$ steps, compared to $22.9$ for the experiments generated by $C_\phi$.
To rule out a potential unfair bias due to different runtimes, Figure~\ref{fig:additional_goal_states} in the Appendix shows an additional baseline of random experiments with an average runtime of $20$ steps, leading to results very similar to those of longer running random experiments.
If we remove the intrinsic adversarial reward, the controller is left only with the external reward.
This means that there is no $\text{bce}$ term in Equation~\ref{eq:loss_c}.
It is not surprising that in this setting, $C_\phi$ fails to generate experiments that discover goal states, since the gradient of $\mathcal{L}_C$ contains no information about the sparse goal reward.

Figure~\ref{fig:force_field_runtime} addresses our second and third hypotheses.
$C_\phi$ indeed tends to prolong experiments  as $M_\mathbf{w}$ has been trained on more experiments, even though experiments with long runtimes are discouraged through the punitive external reward.
Our explanation for this is that it becomes harder with time for $C_\phi$ to come up with short experiments for which $M_\mathbf{w}$ cannot yet accurately predict the correct results.
This is supported by the fact that the prediction accuracy of $M_\mathbf{w}$ for newly generated experiments goes up. Specifically,
Figure~\ref{fig:force_field_runtime} shows the difference between prediction accuracy of the current $M_\mathbf{w}$ for the newly generated experiment and the expected prediction accuracy the current $M_\mathbf{w}$ for randomly sampled experiments.
This accounts for the general gain of $M_\mathbf{w}$'s prediction accuracy over the course of training.
It can be seen that in the beginning, $C_\phi$ is successful at creating adversarial experiments that surprise $M_\mathbf{w}$.
With time, however, it fails to continue doing so and is forced to create longer experiments to challenge $M_\mathbf{w}$.

\begin{algorithm*}[ht]
  \caption{Adversarial yes/no experiments in a differentiable environment}
  \label{alg:adversarial}
   \textbf{Input}: Randomly initialized differentiable Controller $C_\phi: \text{S} \rightarrow \Theta$, randomly initialized differentiable Model $M_\mathbf{w}: \text{S} \times \Theta \rightarrow \mathbb{R}$, empty experiment memory $\mathcal{D}$, empty state memory $\mathcal{B}$, set of random initial experiments $\mathcal{E}_\text{init}$, Differentiable environment
   
   \textbf{Output}: An experiment memory populated with (formerly) interesting experiments
   
    \begin{algorithmic}[1]
    \FOR {$\theta \in \mathcal{E}_\text{init}$}
        \STATE $s \leftarrow$ current environment state
        \STATE Execute the experiment parametrized by $\theta$ in the environment, obtain binary result $r$
        \STATE Save the tuple $(s, \theta, r)$ to $\mathcal{D}$
        \STATE Save all encountered states during the experiment to $\mathcal{B}$
    \ENDFOR
	\REPEAT
	    \STATE $s \leftarrow$ current environment state
		\STATE $\theta \leftarrow C_\phi(s)$
		\STATE Execute the experiment parametrized by $\theta$ in the environment, obtain binary result $r$
		\STATE Save tuple $(s, \theta, r)$ to $\mathcal{D}$
		\STATE $\hat{s} \leftarrow$ current environment state
		\FOR {some steps}
		    \STATE Sample tuple $(s, \theta, r)$ from $\mathcal{D}$
		    \STATE Update the model using SGD: $\nabla_{\textbf{w}}\text{bce}(M_\mathbf{w}(s, \theta), r)$
	    \ENDFOR
		\FOR {some steps}
		    \STATE Sample starting state $s$ from $\mathcal{B}$
		    \STATE Set environment to state $s$
		    \STATE Execute the experiment parametrized by $C_\phi(s)$ in the environment, obtain continuous result $\tilde{r}$ and external reward $R_e$
		    \STATE Update the controller using SGD: $\nabla_{\phi}\big( -\text{bce}(M_\mathbf{w}(s, C_\phi(s)), \tilde{r}) - R_e \big)$
	    \ENDFOR
	    \STATE Set environment to state $\hat{s}$
    \UNTIL{no more interesting experiments are found}
    \end{algorithmic}
\end{algorithm*}

\subsection{Pure RNN Thought Experiments}
\label{sec:pure_thought}

The previous experimental setup uses feedforward NNs as experiments and an intrinsic reward function that is differentiable with respect to the controller's weights.
This section investigates a complementary 
setup: interesting pure thought experiments (with no environment interactions) are generated in the form of RNNs without any inputs, driven by an intrinsic curiosity reward based on information gain which we treat as non-differentiable. 

\subsubsection{Method}

In many ways, this new setup (depicted in Figure~\ref{fig:rnn_schematic} and described in Algorithm~\ref{alg:pure_thought} in the Appendix) is similar to the one presented in Section~\ref{sec:differentiable_env}.
In what follows, we highlight the important differences.

An experiment $E_\theta$ is an RNN of the form $(h_{t+1}, r_{t+1}, \gamma_{t+1}) = E_\theta(h_t) $, where $h_t$ is the hidden state vector, $r_t \in \{0, 1\}$ is the binary result at experiment time step $t$, and $\gamma_t \in [0, 1]$ is the HALT unit.
The result $r$ of $E_\theta$ is the $r_t$ for the experiment step $t$ where $\gamma_t$ first is larger than $0.5$.

Since there is no external environment and the experiments are independent of each other, the model $M_\mathbf{w}$ is again a simple MLP with parameters $\mathbf{w}$.
It takes only the experiment parameters $\theta$ as input and makes a result prediction $\hat{o} = M_\mathbf{w}(\theta), \hat{o} \in [0, 1]$.

As mentioned above, here we treat the intrinsic reward signal as non-differentiable. 
This means that---in contrast to the method presented in Section~\ref{sec:differentiable_env}---the controller cannot receive information about $M_\mathbf{w}$ from gradients that are backpropagated through the model.
Instead, it has to infer the learning behavior of $M_\mathbf{w}$ from the history $\omega$ of previous experiments and intrinsic rewards to come up with new surprising experiments.
The controller $C_\phi$ is now an LSTM that is trained by DDPG \cite{lillicrap2015continuous} and generates new experiments solely based on the history of past experiments: $C_\phi(\omega) = \theta$.
The history $\omega$ is a sequence of tuples $(\theta_i, r_i, R_i)$, where $i = 1, 2, \ldots$ is the index of the experiment. 
It contains experiments up to the last one that has been executed.
More details can be found in Appendix~\ref{app:pure_thought}.

For these pure thought experiments, we use a reward based on information gain.
Let $\mathbf{w}$ be $M$'s weights before training for a fixed number of SGD steps on data that includes the newly generated experiment $\theta$ that has just been added to the memory $\mathcal{D}$, and $\mathbf{w}^*$ the weights after training.
Then, the information gain reward associated with experiment $\theta$ is
\begin{equation}
\label{eq:inf_gain}
    R_{IG}(\theta, \mathbf{w}, \mathbf{w}^*) = \frac{1}{|\mathcal{D}|} \sum_{\tilde{\theta} \in \mathcal{D}} D_{KL}(M_{\mathbf{w}^*}(\tilde{\theta}) || M_\mathbf{w}(\tilde{\theta})), 
\end{equation}
where we interpret the output of the model as a Bernoulli distribution.

\begin{figure}[t]
    \centering
    \begin{minipage}{0.45\textwidth}
        \centering
        \includegraphics[width=0.9\textwidth]{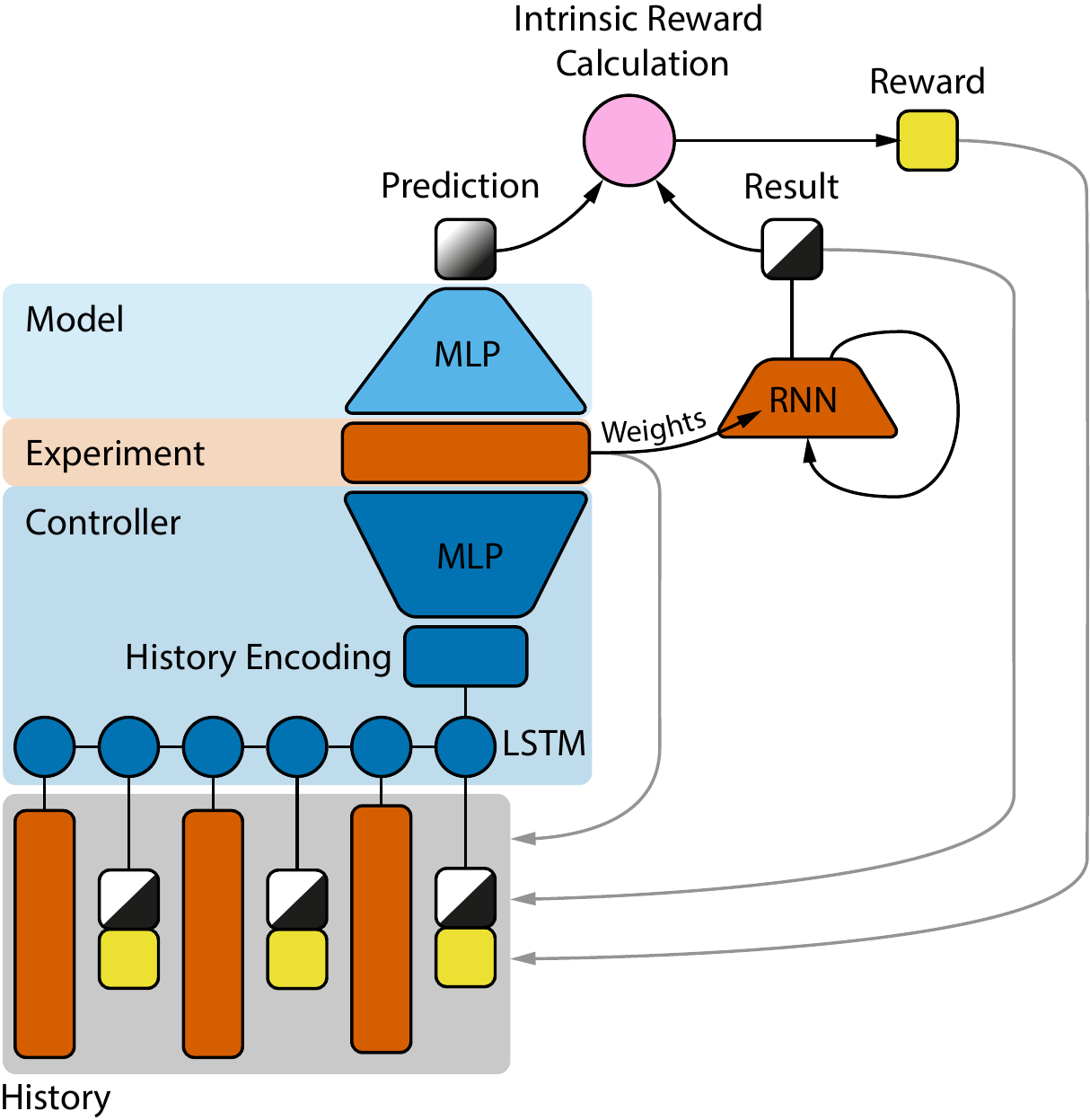} 
        \caption{\textbf{Generating abstract thought experiments encoded as RNNs.} 
        The model is trained to predict the results of previous experiments.
        The controller generates new interesting thought experiments (without environment interactions) based on the history of previous experiments and their results and rewards. 
        } 
        \label{fig:rnn_schematic}
    \end{minipage}\hfill
    \begin{minipage}{0.5\textwidth}
        \centering
        \raisebox{-5.1cm}{\includegraphics[width=1.0\textwidth]{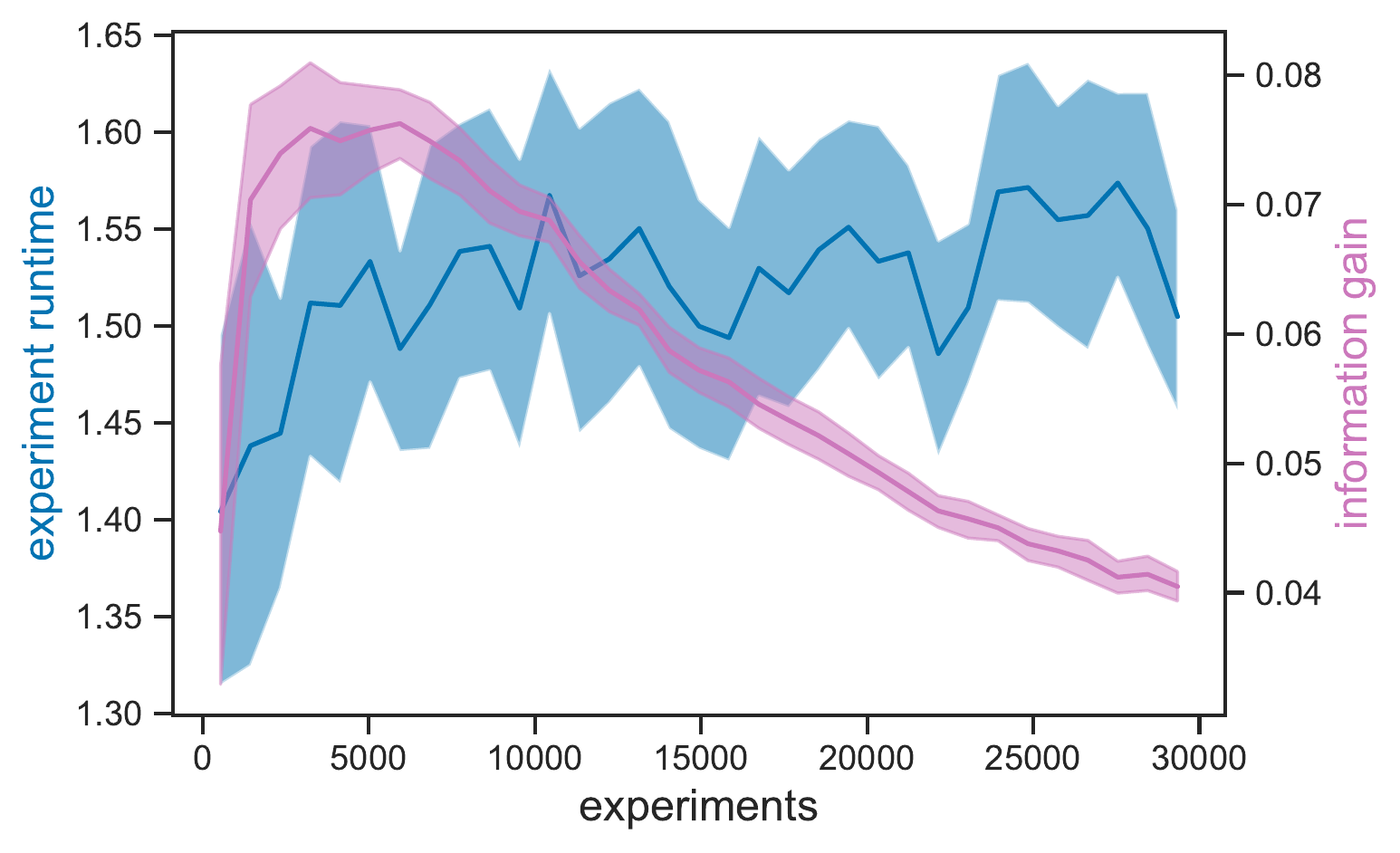}}
        \caption{\textbf{Empirical results for pure thought experiments encoded as RNNs.} Blue: the average runtime of each experiment generated by $C_\phi$. Purple: information gain reward (Equation~\ref{eq:inf_gain}) for $C_\phi$ associated with each experiment. Mean with bootstrapped 95\% confidence intervals across 20 seeds.}
        \label{fig:rnn_results}
    \end{minipage}
\end{figure}

\subsubsection{Results and Discussion}

Figure~\ref{fig:rnn_results} shows the information gain reward associated with each new experiment that $C_\phi$ generates.
We observe that, after a short initial phase, the intrinsic information gain reward steadily declines. 
This is similar to what we observe for the prediction accuracy in section~\ref{sec:force_results}: it becomes harder for the controller to generate experiments that surprise the model.
It should be mentioned that this is a natural effect, since---as the model is trained on more and more experiments---every new additional experiment contributes on average less to the model's change during training, and thus is associated with less information gain reward.

An interesting, albeit minor, effect shown in Figure~\ref{fig:rnn_results} is that also in this setup, the average runtime of the generated experiments increases slightly over time, even though there is no negative reward for longer thought experiments.
For shorter experiments, however, it is apparently easier for the model to learn to predict the results.
Hence, at least in the beginning, they yield more learning progress and more information gain.
Later, however, longer experiments become more interesting.

In comparison to the experiments generated in Section~\ref{sec:differentiable_env}, the present ones have a much shorter runtime.
This is a side-effect of the experiments being RNNs with a HALT unit; for randomly initialized experiments, the average runtime is approximately $1.6$ steps.

\section{Conclusion and Future Work}

We extended the neural Controller-Model (CM) framework through the notion of arbitrary self-invented computational experiments with binary outcomes:
experimental protocols are essentially programs interacting with the environment, encoded as the weight matrices of RNNs generated by the controller.
The model has to predict the outcome of an experiment based solely on the experiment's parameters.
By creating experiments whose outcomes surprise the model, the controller curiously explores its environment and what can be done in it.
Such a system is analogous to a scientist who designs experiments to gain insights about the physical world.
However, an experiment does not necessarily involve actions taken in the environment: it may be  a pure thought experiment akin to those of mathematicians.

We provide an empirical evaluation of two simple instances of such systems, focusing on different and complementary aspects of the idea.
In the first setup, we show that self-invented abstract experiments encoded as feedforward networks interacting with a continuous control environment facilitate the discovery of rewarding goal states.
Furthermore, we see that over time the controller is forced to create longer experiments  (even though this is associated with a larger negative external reward) as short experiments start failing to surprise the model.

In the second setup, the controller generates pure abstract thought experiments in the form of RNNs.
We observe that over time, newly generated experiments result in less intrinsic information gain reward.
Again, later experiments tend to have slightly longer runtime.
We hypothesize that this is because simple experiments initially lead to a lot of information gain per time interval, but later do not provide much insight anymore. 

These two empirical setups should be seen as initial steps towards more capable systems such as the one proposed in Section~\ref{exabs}.
Scaling these methods to more complex environments and the generation of more sophisticated experiments, however, is not without challenges.
Direct generation and interpretation of NN weights may not be very effective for large and deep networks.
Previous work~\cite{faccio2022goal} already combined hypernetworks~\cite{ha2016hypernetworks} and policy fingerprinting~\cite{harb2020policy, faccio2022general} to generate and evaluate policies.
Similar innovations will facilitate the generation of abstract self-invented experiments beyond the small scale setups presented in this paper. 

\section{Acknowledgments} 
\label{ack}

We are grateful to our friends for useful comments. 
This work was supported in part by a European Research Council Advanced Grant (no: 742870).

\bibliography{main.bib}
\bibliographystyle{abbrv}

\newpage
\appendix

\section{Experiments in the Force Field Environment}
\label{app:environment}

The force field of the environment is based on a 2D grid of randomly sampled force vectors.
To get a continuous force field, bicubic interpolation between the vectors of the grid is used.
Hence, the resolution of the grid influences the complexity of the force field (higher resolution $\rightarrow$ more intricate force field).
In all experiments, the grid resolution is sampled uniformly from $\{(3, 3), (5,5), (7,7)\}$.
The random seed of each run affects both the force field and the position of the goal state.
This means that every run has its own unique environment.

\subsection{Experiment Execution}
\label{app:experiments}

Let $\hat{r}_t \in [0, 1]$ be the value of the result node at step $t$ of the experiment whose
runtime is determined by the parameter $\tau \in [0, 100]$.
The maximum runtime is fixed to 100 steps.
A distribution over experiment steps $t$ is defined by $\tau$ as follows:
$
p_\tau(t) = \frac{\exp(-0.5 (t - \tau)^2)}{\sum_{u=1}^{100} \exp(-0.5 (u - \tau)^2)}.
$

The continuous result of the experiment is the expectation of the result unit over this distribution:
$\tilde{r} = \mathbb{E}_{t \sim p_\tau} \hat{r}_t$. 
The binary result of the experiment $r$ is the boolean value $\tilde{r} > 0.5$.

\subsection{Hyperparameters for the Force Field Experiments}
\label{app:hyperparams_force_field}

Table~\ref{table:hyperparams} shows the hyperparameters for Algorithm~\ref{alg:adversarial}.
The output nodes of $C_\phi$ that generate the parameters $\psi$ of the experiment network have a $\text{tanh}$ output nonlinearity and are then scaled to the predefined range.
The output node that generates $\tau$ is clipped to the range $[0, 100]$.

The experiment parameters for random baselines are generated as $\psi = 2 \, \text{tanh}(v)$, where $v \sim \mathcal{N}(0, 4 I)$.
The runtime parameter $\tau$ is sampled uniformly from the allowed range.
The hyperparameters for the model are the same as in Table~\ref{table:hyperparams}.
The baseline with only external reward also uses the hyperparameters of Table~\ref{table:hyperparams}.
The difference is that in this setting, the loss of the $C_\phi$ is simply $\mathcal{L}_C = \mathbb{E}_{s \sim \mathcal{B}} [- R(C_\phi(s), s)]$ instead of the one defined in Equation~\ref{eq:loss_c}.

\begin{table}[h]
\centering
\begin{tabular}{lr}
\textbf{Hyperparameter}                     & \textbf{Value}           \\ \hline
hidden layers $M_\mathbf{w}$                & {[}128, 128, 128, 128{]} \\
hidden layers $C_\phi$                      & {[}128, 128, 128, 128{]} \\
training steps per iteration $M_\mathbf{w}$ & 100                      \\
training steps per iteration $C_\phi$       & 100                      \\
learning rate $M_\mathbf{w}$                & 0.0003                   \\
learning rate $C_\phi$                      & 0.0003                   \\
weight decay $M_\mathbf{w}$                 & 0.01                     \\
weight decay $C_\phi$                       & 0.01                     \\
experiment parameter range                  & {[}-2, 2{]}              \\
noise input nodes $C_\phi$                  & 8                        \\
environment grid resolutions                & [(3, 3), (5, 5), (7, 7)] \\
number of iterations                        & 1000                     \\
number of initial experiments in $\mathcal{E}_\text{init}$     & 100          \\
\end{tabular}
\caption{Hyperparameters for Algorithm~\ref{alg:adversarial}}
\label{table:hyperparams}
\end{table}

\subsection{Additional Results}
\label{app:additional_results}

\begin{figure}[ht]
     \centering
     \begin{subfigure}[t]{0.45\textwidth}
         \centering
         \includegraphics[width=\textwidth]{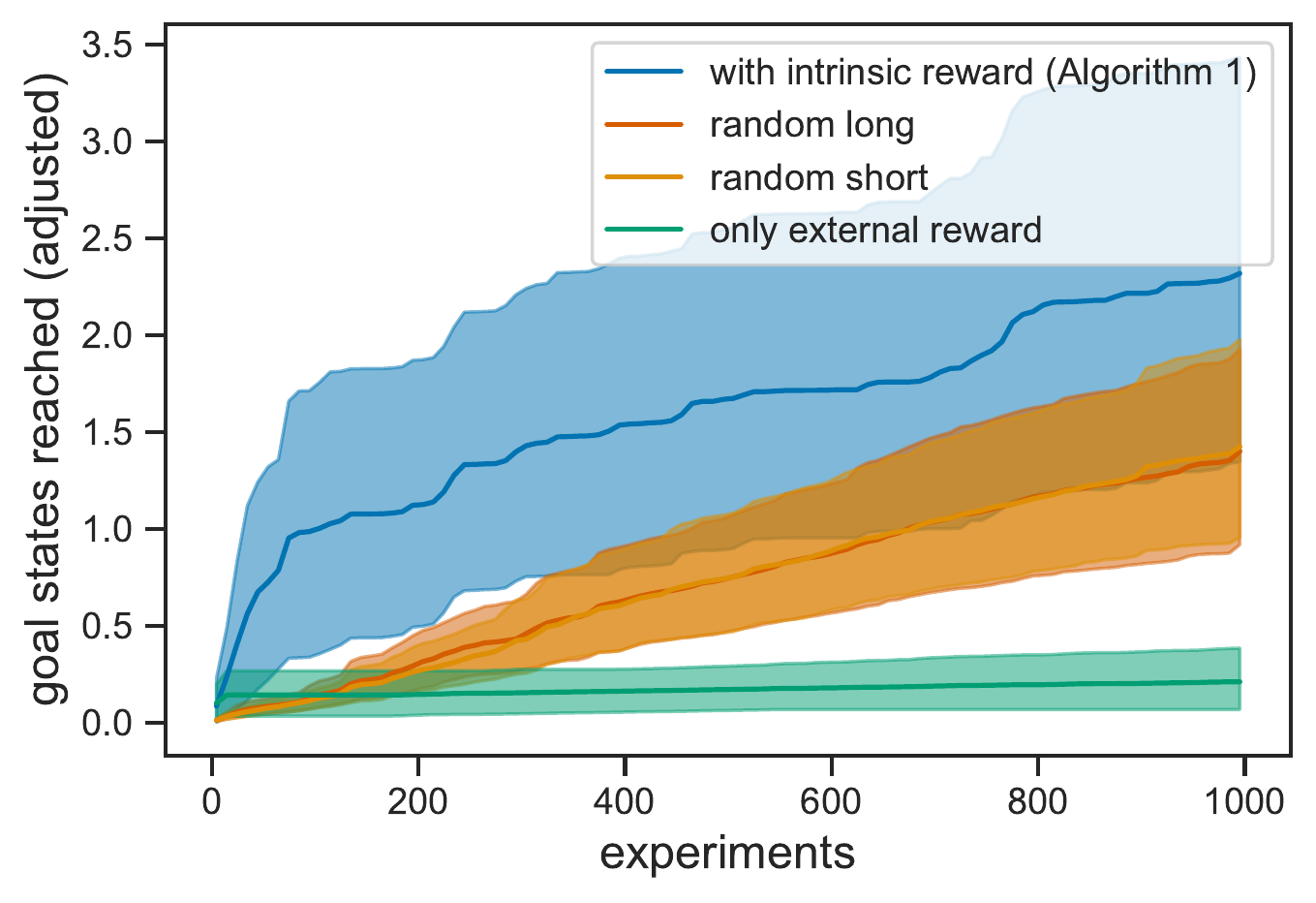}
     \end{subfigure}
     \caption{Similar to Figure~\ref{fig:force_field_goal_states}, but with an additional baseline of short random experiments with an average runtime of 20 steps.}
     \label{fig:additional_goal_states}
     \hfill
\end{figure}

To account for a potential bias due to experimental runtime, Figure~\ref{fig:additional_goal_states} shows the adjusted number of goal states for a baseline of shorter random experiments.

\section{Pure Thought Experiments}
\label{app:pure_thought}

Algorithm~\ref{alg:pure_thought} summarizes the method described in Section~\ref{sec:pure_thought}.
In this setup, the model $M_\mathbf{w}$ is trained to minimize the following loss:

\begin{equation}
\label{eq:loss_m_pure_thought}
\mathcal{L}_M = \mathbb{E}_{(\theta, r) \sim \mathcal{D}}[\text{bce}(M_\mathbf{w}(\theta), r)].
\end{equation}

Efficient approximation of the policy gradients for the controller is achieved through an actor-critic method, specifically DDPG \cite{lillicrap2015continuous}.
The controller $C_\phi$ has an additional LSTM encoder that generates a vector-sized representation of the history $\omega$ of previous experiments, their results and the reward associated with them.
The actor is an MLP that receives as input the history representation created by the LSTM and generates the weights of an experiment RNN, whereas the critic receives both a history representation and experiment weights as input, and outputs a scalar reward estimation. 
Actor and critic share the same LSTM history encoder and take alternating gradient descent steps during training.
The input to the LSTM history encoder is the sequence $\omega$ of the last $1000$ that have been executed.

The experiment RNNs $E_\theta$ used in this empirical evaluation have $3$ hidden units and no inputs. 
The initial hidden state $h_0$ is treated as part of the parameters $\theta$ and is thus also generated by $C_\phi$.
Random experiments are sampled the same way as described in Section~\ref{app:hyperparams_force_field}.
All other hyperparameters are listed in Table~\ref{table:hyperparams_pure_thought}.

\begin{algorithm*}[h]
  \caption{Pure thought experiments encoded by RNNs}
  \label{alg:pure_thought}
   \textbf{Input}: Randomly initialized differentiable Controller $C_\phi: \Omega \rightarrow \Theta$, where $\Omega$ is the set of sequences of the form $(\theta_i, r_i, R_i, \theta_{i+1}, r_{i+1}, R_{i+1}, \ldots)$, randomly initialized differentiable Model $M_\mathbf{w}: \Theta \rightarrow \mathbb{R}$, empty sequential experiment memory $\mathcal{D}$, set of random initial experiments $\mathcal{E}_\text{init}$ 
   
   \textbf{Output}: An experiment memory populated with (formerly) interesting pure thought experiments 
   
    \begin{algorithmic}[1]
    \FOR {$\theta \in \mathcal{E}_\text{init}$}
        \STATE Execute the RNN thought experiment parametrized by $\theta$, obtain binary result $r$
        \STATE Save the tuple $(\theta, r)$ to $\mathcal{D}$
        \STATE Train $M_\mathbf{w}$ on data from $\mathcal{D}$ for a fixed number of steps minimizing Equation~\ref{eq:loss_m_pure_thought} to obtain updated weights $\mathbf{w}^*$
        \STATE Calculate the intrinsic reward $R_i = R_{IG}(\theta, \mathbf{w}, \mathbf{w}^*)$ (Equation~\ref{eq:inf_gain})
        \STATE $\mathbf{w} \leftarrow \mathbf{w}^*$
        \STATE Save $R_i$ to $\mathcal{D}$
    \ENDFOR
	\REPEAT
	    \STATE $\omega \leftarrow$ sequence of the last experiments from $\mathcal{D}$
		\STATE $\theta \leftarrow C_\phi(\omega)$
        \STATE Execute the RNN thought experiment parametrized by $\theta$, obtain binary result $r$
        \STATE Train $M_\mathbf{w}$ on data from $\mathcal{D}$ for a fixed number of steps to obtain updated weights $\mathbf{w}^*$
        \STATE Calculate the intrinsic reward $R_i = R_{IG}(\theta, \mathbf{w}, \mathbf{w}^*)$
        \STATE $\mathbf{w} \leftarrow \mathbf{w}^*$
        \STATE Save $R_i$ to $\mathcal{D}$
        \STATE Train $C_\phi$ for a fixed number of steps with DDPG to maximize the expected intrinsic reward
    \UNTIL{no more interesting experiments are found}
    \end{algorithmic}
\end{algorithm*}

\begin{table}[h]
\centering
\begin{tabular}{lr}
\textbf{Hyperparameter}                     & \textbf{Value}           \\ \hline
hidden layers $M_\mathbf{w}$                & {[}128, 128, 128, 128{]} \\
hidden layers $C_\phi$ LSTM                 & {[}64{]}                 \\
hidden layers $C_\phi$ MLP                  & {[}128, 128, 128, 128{]} \\
training steps per iteration $M_\mathbf{w}$ & 50                       \\
training steps per iteration $C_\phi$       & 10                       \\
learning rate $M_\mathbf{w}$                & 0.0001                   \\
learning rate $C_\phi$                      & 0.0001                   \\
weight decay $M_\mathbf{w}$                 & 0.01                     \\
weight decay $C_\phi$                       & 0.01                     \\
experiment parameter range                  & {[}-3, 3{]}              \\
number of iterations                        & 30000                    \\
number of initial experiments in $\mathcal{E}_\text{init}$     & 100          \\
\end{tabular}
\caption{Hyperparameters for Algorithm~\ref{alg:pure_thought}}
\label{table:hyperparams_pure_thought}
\end{table}

\end{document}